\documentclass[letterpaper, 10 pt, journal, twoside]{IEEEtran}

\IEEEoverridecommandlockouts                              % This command is only needed if 
\title{
Bayesian Gaussian Mixture Model\\for Robotic Policy Imitation
}

\author{Emmanuel Pignat, and Sylvain Calinon%
	\thanks{Manuscript received: February, 24, 2019; Revised June, 1, 2019;
		Accepted July, 12, 2019.}%Use only for final RAL version
	\thanks{This paper was recommended for publication by Editor Dongheui Lee upon
		evaluation of the Associate Editor and Reviewers' comments. The research leading to these results has received funding from the European Commission's Horizon 2020 Programme through the MEMMO Project (Memory of Motion, \href{http://www.memmo-project.eu/}{http://www.memmo-project.eu/}, grant agreement 780684) and CoLLaboratE project (\href{https://collaborate-project.eu/}{https://collaborate-project.eu/}, grant agreement 820767).}%Use only for final
	\thanks{The authors are with the Idiap Research Institute, Martigny, Switzerland. {\tt\small  emmanuel.pignat@idiap.ch, sylvain.calinon@idiap.ch}.}%
	\thanks{Digital Object Identifier (DOI): see top of this page.}
}
\usepackage{graphicx,amsmath,amssymb,array,fancyhdr,color,sidecap,paralist,eurosym,nicefrac,lastpage}%bibspacingmathtools,scalefnt,nicefrac
\usepackage[colorlinks=false, pdfborder={0 0 0}]{hyperref}
\usepackage{tikz}
\usepackage{bm}
\usepackage{booktabs}
\usepackage[percent]{overpic}
\usepackage{hyphenat}
\usepackage{color}

\newcommand{\ee}{^{\mathrm{ee}}}
\newcommand{\nbprod}{M}

\newcommand{\outgin}{\bm{y}|\bm{x}}
\newcommand{\outout}{\bm{y}}
\newcommand{\outin}{\bm{y}\bm{x}}
\newcommand{\inout}{\bm{x}\bm{y}}
\newcommand{\inin}{\bm{x}}
\newcommand{\new}[1]{#1}
\newcommand{\trsp}{{\scriptscriptstyle\top}}

\begin{document}

\bibliographystyle{IEEEtran}

\markboth{IEEE Robotics and Automation Letters. Preprint Version. Accepted July, 2019}
{Pignat\MakeLowercase{\textit{et al.}}: Bayesian Gaussian Mixture Model for Robotic Policy Imitation}  

\maketitle
%\thispagestyle{empty}
%\pagestyle{empty}

%%%%%%%%%%%%%%%%%%%%%%%%%%%%%%%%%%%%%%%%%%%%%%%%%%%%%%%%%%%%%%%%%%%%%%%%%%%%%%%%
\begin{abstract}
A common approach to learn robotic skills is to imitate a demonstrated policy.
Due to the compounding of small errors and perturbations, this approach may let the robot leave the states in which the demonstrations were provided.
This requires the consideration of additional strategies to guarantee that the robot will behave appropriately when facing unknown states.
We propose to use a Bayesian method to quantify the action uncertainty at each state. 
The proposed Bayesian method is simple to set up, computationally efficient, and can adapt to a wide range of problems.
Our approach exploits the estimated uncertainty to fuse the imitation policy with additional policies.
It is validated on a Panda robot with the imitation of three manipulation tasks in the continuous domain using different control input/state pairs. 
\end{abstract}

%

%\begin{keywords} % had to change to RAL official keywords
%	Learning by Demonstration, Learning and Adaptive Systems, Probability and Statistical Methods
%\end{keywords} 

%%%%%%%%%%%%%%%%%%%%%%%%%%%%%%%%%%%%%%%%%%%%%%%%%%%%%%%%%%%%%%%%%%%%%%%%%%%%%%%%
\section{Introduction}

Many learning modalities exist to acquire robot manipulation tasks.
Reward-based methods, such as optimal control (OC) or reinforcement learning (RL), either require accurate models or a large number of samples.
An appealing approach is behavior cloning (or policy imitation), where the robot learns to imitate a policy that consists of a conditional model $p(\bm{u}_t | \bm{x}_t)$ retrieving a control command $\bm{u}_t$ at state $\bm{x}_t$.
Due to modeling errors, perturbations or different initial conditions, executing such policy can quickly lead the robot far from the distribution of states visited during the learning phase.
This problem is often referred to as the distributional shift \cite{ross2011reduction}. 
When applied to a real system, the actions can therefore be dangerous and lead to catastrophic consequences.
Many approaches, such as \cite{ross2011reduction}, have been addressing this problem in the general case. 
% TODO to cite or note in the following sentence
A subset of these approaches focus on learning manipulation tasks from a small set of demonstrations \cite{hersch2008dynamical,khansari2011learning,paraschos2013probabilistic}. 
In order to guarantee safe actions, these techniques typically add constraints to the policy, by introducing time-dependence structures or by developing hybrid, less general approaches. % which we target in a separate sentence ?
% TODO should I talk about heavy computation % TODO REORGANISE PARAGRAPH
We propose to keep the flexibility of policy imitation without constraining heavily the policy to be learned by relying on Bayesian models, providing uncertainty quantification measures. %% TODO DONE REVIEW explain better what comes next
%"Uncertainty quantification is very important in behaviour cloning due
%to the problems mentioned before." is vague. Authors should highlight
%the advantages of such an approach more.
Uncertainty quantification can be exploited in various ways, but such capability often comes at the expense of being computationally demanding. 
In this work, we propose a computationally efficient and simple Bayesian model that can be used for policy imitation. % or "model for policy imitation"
It allows active learning or fusion of policies, which will be detailed in Sec.~\ref{sec:policy_fusion}. %TODO Sec or in the Section, ...
The flexibility of the proposed model can be exploited in wide-ranging data problems, in which the robot can start learning from a small set of data, without limiting its capability to increase the complexity of the task when more data become available.

%We propose to use Bayesian models, allows to quantify the uncertainty about the policy, fuse multiple policies in a smart way or toggle active learning.
%Bayesian models, intractable, we propose to use conditioning in Bayesian Gaussian mixture model. Very efficient training, below noticeable time, very interactive teaching.
%\begin{align*}
% p(\bm{u}_t | \bm{x}_t)
%\end{align*}

\begin{figure}
	\centering
	\includegraphics[width=.9\linewidth]{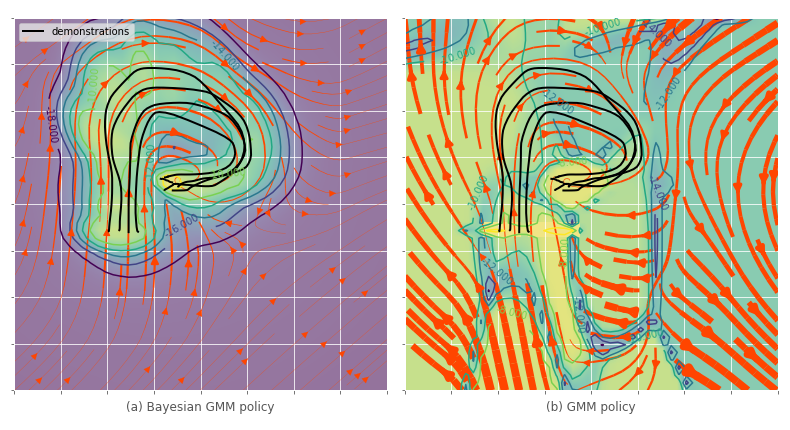}
	\caption{Comparison of Bayesian and non-Bayesian GMM conditional distributions. The flow field represents the expectation of $p(\bm{u}|\bm{x}, \bm{X}, \bm{U})$ and the colormap its entropy $\mathcal{H}(p(\bm{u}|\bm{x}, \bm{X}, \bm{U}))$. Yellow color is for lower entropy, meaning higher certainty. The black lines are the demonstrations $\bm{X}$ %$p(\dot{\bm{x}}|\bm{x})$
		\textit{(a) Bayesian model:} Certainty is localized in the vicinity of the demonstrations. The policy retrieved further away can result in poor generalization, but the system is aware of this through the uncertainty estimate. \textit{(b) Non-Bayesian model:} The entropy only relates to the variations of the demonstrated policy instead of a Bayesian uncertainty. 
}
	\label{fig:gmrvsbayesian}
\end{figure}

For didactic and visualization purposes, we will consider throughout the article a simple 2D velocity controlled system with state as position (see e.g., Fig.~\ref{fig:gmrvsbayesian}). However, the approach is developed for higher dimensional systems, which will be demonstrated in Sec.~\ref{sec:experiments} with velocity and force control of a 7-axis manipulator.
% TODO cite as in \cite{khansari2011learning}.
%It allows us to visualize the policy in one image, using flow-fields. % TODO or policy, for a part of the state space

\section{Related work}
%% General policy
%Compounding error, one timestep prediction, distributional shift (move to another distribution of where it was trained), dangerous states \cite{ross2011reduction}

Several works have tackled the distributional shift problem in the general case. % TODO only let distributional shift if mentioned above
In \cite{ross2011reduction}, Ross et al.\ used a combination of expert and learner policy. % TODO can we cite at the beginnning of a sentence ?
%iid assumption \cite{ross2011reduction} dagger \cite{ross2010efficient} ask for corrections at new states (dangerous to visit unknown states, tedious learning)
%\cite{menda2017dropoutdagger} use dropout as Bayesian %\cite{gal2016dropout} %\cite{cronrathbagger}
In \cite{laskey2017dart}, perturbations are added in order to force the expert to demonstrate recovery strategies, resulting in a more robust policy. % TODO remove "from them", TODO !!! is also applied to robotic manipulation, maybe show that is compatible with our approach
%the injecting noise to force the demonstrator to show how to recover from perturbation

% TODO DONE REVIEW : GMM conditionin -> be more explicit
%Sec. II, second paragraph: "More closely related to our manipulation
%tasks, Khansari- Zadeh et al. have also used GMM conditioning to learn
%a policy on the form ... [3]”. What do the authors mean by "GMM
%conditioning"?	This concept keeps being repeated further in the text,
%but is never explicitly explained. In the provided citations ([2] and
%[3]), such concept does not appear either.

%%TODO REVIEW : be more clear that DMP escapes unstability problems by the structure they impose on the task, which limits its applicability
%1. What is the motivation for using a Bayesian model? DMPs can follow
%demonstrations arbitrarily well, and generalize to disturbances. Can
%the authors compare to DMPs, to provide some context where DMPs might
%fail, but the Bayesian framework succeeds?

More closely related to our manipulation tasks, Khansari-Zadeh et al.\ have used a similar approach to ours to learn a policy of the form $\bm{\dot{x}} = f(\bm{x})$ \cite{khansari2011learning}.
They used conditional distributions $p(\bm{\dot{x}}|\bm{x})$ in a joint model of $p(\bm{x}, \bm{\dot{x}})$ represented as a Gaussian mixture model (GMM).
A structure was imposed on the parameters to guarantee asymptotic convergence. %but limit the variety of tasks that can be learned. 
As an alternative, we propose to exploit the uncertainty information of a Bayesian GMM, resulting in a less constrained approach that can scale easily to higher dimensions.

Dynamical movement primitives (DMP) is a popular approach that combines a stable controller (spring-damper system) with non-linear forcing terms decaying over time through the use of a phase variable, ensuring convergence at the end of the motion \cite{schaal2005learning}.
A similar approach is used in \cite{hersch2008dynamical} where the non-linear part is encoded using conditional distributions in GMM.
Due to their underlying time dependence, these approaches are often limited to either point-to-point or cyclic motions of known period, with limited temporal and spatial robustness to perturbations.

Another approach to avoid these problems is to model distributions of states or trajectories instead of policies \cite{calinon2014task,paraschos2013probabilistic}. 
However, these techniques are limited to the imitation of trajectories and cannot handle easily misaligned or partial demonstrations of movements.
%They are closer to inverse optimal control and typically rely on a model of the system for synthesis.

Inverse optimal control \cite{finn2016guided}, which tries to find the objective minimized by the demonstrations, is another direction to increase robustness and generalization.
It often requires more training data and is computationally heavy.

If a reward function for the task is accessible, an interesting alternative is to combine the proposed policy imitation strategy with reinforcement learning \cite{rajeswaran2017learning,nair2018overcoming}, where the imitation loss can reduce the exploration phase while the reinforcement learning overcomes the limits of imitation.
%% robotic manipulation tasks, small number of demonstrations
%solution 3:
%\paragraph{Time dependent}
% time-dependent ensure convergence at the end
%Need to align trajectories, mostly open loop \cite{schaal2005learning} % dmp time-dependent clock, heuristic
% dx = f(t) or ddx = f(x, dx) using GMR
%\cite{paraschos2013probabilistic} % special treatmetns with loops
%TODO cite it later \cite{paraschos2015model} time dependent and additional controller to converge back to demonstration with switching rule
%\paragraph{Policy imitation}
%hard constraints
%\cite{khansari2011learning} % unique attractors, no limit cycles, need to know the system we are controlling, velocity controlled
%\cite{schaal2002scalable} % check because zero outside area, emphasis on online incremental
%\paragraph{Reinforcement learning and inverse}
%% force policy

% bayesian
%\paragraph{Other}
%\cite{wang2006gaussian} predicting time-series, predicting variance
%online learning 

%% TODO REVIEW : we expect the demonstrations to be far from optimal, distribution of close to optimal
%% a weighting might be feasible, failed demonstrations would need to be taken as example of what not to do
%% PROBLEM, how to assess from quality of demonstrations (demonstrator self evaluation), how to know where it fails =

%2. What if the demonstrations are of varying quality (not just
%multi-modal)? Can the framework generalize to failed demonstrations?
%Can there be a weighing based on the quality of the demonstration?
%Maybe the authors can add a discussion about this.

\section{Bayesian Gaussian mixture model conditioning}
\label{sec:gmm}
% TODO also tell that it is fully derivable
%\subsection{Motivation}
%Neural network no uncertainty. Bayesian yes but with higher cost. Is efficient Our model alleviate
Numerous regression techniques exist and have been applied to robotics. In this section, we derive a Bayesian version of Gaussian mixture conditioning \cite{sung2004gaussian}, which was already used in \cite{hersch2008dynamical,khansari2011learning}. Besides uncertainty quantification, we start by listing the characteristics required in our tasks to motivate the need for a different approach.  

% A lot of interesting features come at a very low computation cost and with very few parameters to tune.

% TODO DONE REVIEW : explain why multimodal is important, e.g. obstacles
%In Sec. III, the authors state that using Gaussian mixture models
%(GMMs) is more beneficial than	GPs and LWPR when learning policies
%from human demonstrations, because these tend to be clearly multimodal.
%Could the authors provide a citation/evidence for this claim?
%Human-demonstrated policies being multimodal seems to be an important
%claim that the authors use as support for the derived framework,
%therefore, I suggest to support it.

\begin{figure}
	\centering
	\includegraphics[width=.9\linewidth]{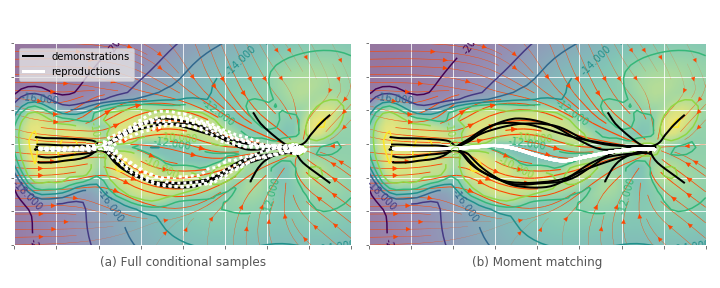}
	\caption{\new{Advantages of encoding multimodal policy $p(\bm{u}|\bm{x})$. The demonstrations show how to reach a point while avoiding an obstacle. At a given position, the distribution of velocity is bimodal, indicating to go either upwards or downwards. \textit{(a)} Multiple reproductions executed by sampling from an expressive multimodal policy $p(\bm{u}|\bm{x})$ as in \eqref{equ:posterior_cond_pred}. \textit{(b)} Reproduction with a unimodal policy, where moment matching was applied to \eqref{equ:posterior_cond_pred}. The average command indicates to go straight, which results in a failure.}}
	\label{fig:mmvsmm}
\end{figure}

\paragraph{Multimodal conditional} Policies from human demonstrations are never optimal but exhibit some stochasticity. In some tasks, for example implying obstacle avoidance, there could be multiple clearly separated paths. Fig.~\ref{fig:mmvsmm} illustrates this idea. Our approach should be able to encode arbitrarily complex conditional distributions. Existing approaches such as locally weighted projection regression (LWPR) \cite{schaal2002scalable} or Gaussian process regression (GPR) \cite{rasmussen2004gaussian} only model unimodal conditional distribution. In its original form, GPR assumes homoscedasticity (constant covariance over the state).

\paragraph{Efficient and robust computation} Most Bayesian models can be computationally demanding for both learning and prediction. For fast and interactive learning, we seek to keep a low training time (below 5 sec for the tasks shown in the experiments).
For reactivity and stability, the prediction should be below 1 ms.
We seek to avoid difficult model selection/hyperparameters tuning, for the approach to be applied in a wide range of tasks, controllers and robotic platforms.
%Besides time, we also want to avoid the pitfalls of optimization-based training. 
%They require to approximate intractable integrals, both at learning and prediction time.

%\paragraph{Uncertainty quantification} The method should be able 
\paragraph{Wide-ranging data} The approach should be able to encode very simple policies with a few datapoints and more complex ones when more datapoints are available.

%\begin{itemize}
%	\item \textit{Multimodal conditional :} policy from human demonstrations are never optimal but exhibit some stochasticity. In some tasks (e.g. implying obstacle), the policy can even be clearly multimodal. Arbitrarily complex conditional distributions. Existing approaches as LWPR or GP only model unimodal conditional. In its original form, GP assumes homoscedasticity, meaning constant covariance over the state. 
%	\item \textit{Uncertainty quantification :}
%	\item \textit{Wide-ranging data :}
%	\item \textit{Efficient computation :}
%\end{itemize}

%Generative models are known to be easier and faster to train than conditional models.
Representing a joint distribution as a Gaussian mixture model and computing conditional distribution is simple and fast \cite{sung2004gaussian}, with various applications in robotics \cite{hersch2008dynamical,khansari2011learning}. % DONE TODO check apply to or in
%In this section, we will present the required components to derive its Bayesian version. % TODO maybe change ingredients
Bayesian regression methods approximate the posterior distribution $p(\bm{\theta}|\bm{X}, \bm{Y})$ of the model parameters $\bm{\theta}$, given input $\bm{X}$ and output $\bm{Y}$ datasets.\footnote{Here, we refer to the output as $\bm{y}$, which corresponds to the control command $\bm{u}$ in an imitation problem.} % DONE TODO check input-output dataset
Then, predictions are made by marginalizing over the parameters posterior distribution, in order to propagate model uncertainties to predictions 
\begin{align}
p(\bm{y}|\bm{x}, \bm{X}, \bm{Y}) =
\int_{\bm{\theta}} p(\bm{y}|\bm{x}, \bm{\theta})p(\bm{\theta}|\bm{X}, \bm{Y})d\bm{\theta},
\end{align}
where $\bm{x}$ is a new query. 
This distribution is called posterior predictive. 
Parametric non-Bayesian methods typically rely on a single point estimate of the parameters $\bm{\theta}^\star$ such as maximum likelihood. % TODO single or not
A variety of methods, adapted to the various model, exists for approximating $p(\bm{\theta}|\bm{X}, \bm{Y})$. For example, variational methods, Monte Carlo methods or expectation propagation. % TODO for approximating or to approximate ?
For its scalability and efficiency, we present here a variational method using conjugate priors and mean field approximation, see \cite{bishop2006pattern} for details. % TODO check mean-field or mean field

\subsection{Bayesian analysis of multivariate normal distribution}
\newcommand{\xy}{\bm{z}}
\newcommand{\XY}{\bm{Z}}
We first focus on the Bayesian analysis of multivariate normal distribution (MVN), also detailed in \cite{murphy2007conjugate}, and then treat the mixture case. % TODO maybe reformulate last part of the sentence, in next subsection ?
As a notation, we use $\xy$ to denote $\begin{bmatrix}\bm{x}^\trsp\, \bm{y}^\trsp \end{bmatrix}^\trsp$, a joint observation of input and output. $\XY$ denotes the joint dataset.

\paragraph{Prior}
The conjugate prior of the MVN is the normal-Wishart distribution.
The convenience of a conjugate prior is the closed-form expression of the posterior. %, avoiding problems of approximate techniques. % TODO of or for the posterior, the problems or problems 
The normal-Wishart distribution is a distribution over mean $\bm{\mu}$ and precision matrix $\bm{\Lambda}$, % TODO mean vector ? or mean should I put a comma
\begin{align}
p(\bm{\mu}, \bm{\Lambda}) &= \mathcal{NW}(\bm{\mu}, \bm{\Lambda}|\bm{\mu}_0, \kappa, \nu, \bm{T}),\\
&= \mathcal{N}\big(\bm{\mu}|\bm{\mu}_0, (\kappa\bm{\Lambda})^{-1}\big)\mathcal{W}_{\nu}(\bm{\Lambda}|\bm{T}),
\end{align}
where $\bm{T}$ and $\nu$ are the scale matrix and the degree of freedom of the Wishart distribution, and $\kappa$ is the precision of the mean.

\paragraph{Posterior}
The closed form expression for the posterior is 
\begin{align}
p(\bm{\mu}, \bm{\Lambda}|\bm{X}) &= \mathcal{N}\big(\bm{\mu}|\bm{\mu}_n, (\kappa_n\bm{\Lambda}_n)^{-1}\big)\mathcal{W}_{\nu_n}(\bm{\Lambda}_n|\bm{T}_n),\\
\bm{\mu}_n &= \frac{\kappa\bm{\mu}_0 + n\bar{\xy}}{\kappa + n},\\
\bm{T}_n &= \bm{T} + \bm{S} + \frac{\kappa n}{\kappa + n}(\xy_0 - \bar{\xy}) (\bm{\mu}_0 - \bar{\xy})^\trsp,\\
\bm{S} &= \sum_{i=1}^{n}(\xy_i - \bar{\xy})(\xy_i - \bar{\xy})^\trsp,\\
\nu_n &= \nu + n, \quad
\kappa_n = \kappa + n,
\end{align}
where $n$ is the number of observations and $\bar{\xy}$ is the empirical mean.
As for conjugate priors, the prior distribution can be interpreted as pseudo-observations to which the dataset is added.

% TODO DONE REVIEW : check for posterior predictive or predictive posterior :)
%Sec. III.c): "Posterior predictive."
%I am confused by this title. In the paragraph right above Sec. III.A,
%this is called "predictive posterior". I suggest to use only one of the
%two names, for consistency.

\paragraph{Posterior predictive}
% TODO DONE REVIEW, d dimensionality of data what is d is Eq (9)?
% TODO DONE REVIEW, explain better that it "IS" a T-distribution
% what is the motivation for representing the posterior predictive
%distribution with a t-distribution?
%TODO can we use an abbreviation for multivariate t-distribution ?
Computing the posterior predictive distribution
\begin{align}
p(\xy|\XY) = \int_{\bm{\Lambda}, \bm{\mu}} p(\xy|\bm{\mu}, \bm{\Lambda}) p(\bm{\mu}, \bm{\Lambda}|\XY) d\bm{\Lambda}
 d\bm{\mu},
 \label{equ:mvn_posterior_pred}
\end{align}
yields a multivariate t-distribution with degree of freedom $\nu_n - d + 1$
\begin{align}
p(\xy|\XY) = t_{\nu_n - d + 1}\big( \xy |\, \bm{\mu}_n, \frac{\bm{T}_n(\kappa_n + 1)}{\kappa_n(\nu_n - d + 1)}\big),
\label{equ:mvn_posterior_pred_t}
\end{align}
where $d$ is the dimensionality of $\xy$.
The multivariate t-distribution has heavier tails than the MVN.
The MVN is a special case of this latter, when the degree of freedom parameter tends to infinity, which corresponds to having infinitely many observations. %$nu_n \to\infty$

\paragraph{Conditional posterior predictive}
For our application, we are interested in computing conditional distributions in the joint distribution of our input $\bm{x}$ and output $\bm{y}$. % TODO work on this sentence
We rewrite the result from \eqref{equ:mvn_posterior_pred} as % TODO change present
\begin{align}
p(\bm{x}, \bm{y}\,|\bm{X}, \bm{Y}) = t_{\nu_{\xy}}(\bm{\mu}_{\xy}, \bm{\Sigma}_{\xy}).
\end{align}
Following \cite{roth2012multivariate}, the multivariate t-distribution conditional distribution is also a multivariate t-distribution, 
\begin{align}
p(\bm{y}\,|\bm{x}, \bm{X}, \bm{Y}) = t_{\nu_{\outgin}}(\bm{\mu}_{\outgin}, \bm{\Sigma}_{\outgin}),
\end{align}
with
%\begin{align}
%	\label{mtd_condition_nu}
%%\nu_{\xout|\xin} &= \nu + d_\xin\\
%%\bm{\mu}_{\xout|\xin} &= \bm{\mu}_\xout + \bm{\Sigma}_{\xout\xin}\bm{\Sigma}_{\xin\xin}^{-1}(\bm{x}_\xin-\bm{\mu}_\xin)\\
%%\bm{\Sigma}_{\xout|\xin} &= \frac{\nu + (\bm{x}_\xin-\bm{\mu}_\xin)^\trsp\bm{\Sigma}_{\xin\xin}^{-1}(\bm{x}_\xin-\bm{\mu}_\xin)}{\nu + d_\xin}(\bm{\Sigma}_{\xout\xout}-\bm{\Sigma}_{\xout\xin}\bm{\Sigma}_{\xin\xin}^{-1}\bm{\Sigma}_{\xout\xin}^\trsp)\\
%\nu_{\outgin} &= \nu_{\inout} + d_{\inin},\\
%\bm{\mu}_{\outgin} &= \bm{\mu}_{\outout} + \bm{\Sigma}_{\outin}\bm{\Sigma}_{\inin}^{-1}(\bm{x}-\bm{\mu}_{\inin}),\\
%\bm{\Sigma}_{\outgin} &= \frac{\nu_{\inout} + (\bm{x}-\bm{\mu}_{\inin})^\trsp\bm{\Sigma}_{\inin}^{-1}(\bm{x}-\bm{\mu}_{\inin})}{\nu_{\inout} + d_{\inin}}\notag\\
%\quad\quad&\times(\bm{\Sigma}_{\outout}-\bm{\Sigma}_{\outin}\bm{\Sigma}_{\inin}^{-1}\bm{\Sigma}_{\outin}^\trsp),
%\label{mtd_condition_scale}
%\end{align}
\begin{align}
\label{mtd_condition_nu}
%\nu_{\xout|\xin} &= \nu + d_\xin\\
%\bm{\mu}_{\xout|\xin} &= \bm{\mu}_\xout + \bm{\Sigma}_{\xout\xin}\bm{\Sigma}_{\xin\xin}^{-1}(\bm{x}_\xin-\bm{\mu}_\xin)\\
%\bm{\Sigma}_{\xout|\xin} &= \frac{\nu + (\bm{x}_\xin-\bm{\mu}_\xin)^\trsp\bm{\Sigma}_{\xin\xin}^{-1}(\bm{x}_\xin-\bm{\mu}_\xin)}{\nu + d_\xin}(\bm{\Sigma}_{\xout\xout}-\bm{\Sigma}_{\xout\xin}\bm{\Sigma}_{\xin\xin}^{-1}\bm{\Sigma}_{\xout\xin}^\trsp)\\
\nu_{\outgin} &= \nu_{\inout} + d_{\inin},\\
\bm{\mu}_{\outgin} &= \bm{\mu}_{\outout} + \bm{\Sigma}_{\outin}\bm{\Sigma}_{\inin}^{-1}(\bm{x}-\bm{\mu}_{\inin}),\\
\bm{\Sigma}_{\outgin} &= \frac{\nu_{\inout}\! +\! (\bm{x}\!-\!\bm{\mu}_{\inin})^\trsp\bm{\Sigma}_{\inin}^{-1}(\bm{x}\!-\!\bm{\mu}_{\inin})}{\nu_{\inout} + d_{\inin}}(\bm{\Sigma}_{\outout}\!-\!\bm{\Sigma}_{\outin}\bm{\Sigma}_{\inin}^{-1}\bm{\Sigma}_{\outin}^\trsp),
\label{mtd_condition_scale}
\end{align}
where $d_{\inin}$ is the dimension of $\bm{x}$, with $\bm{\mu}_{\xy}$ and $\bm{\Sigma}_{\xy}$ decomposed as 
\begin{align}
	\bm{\mu}_{\xy} =
	\begin{bmatrix}
	\bm{\mu}_{\inin}\\ 
	\bm{\mu}_{\outout}\\ 
	\end{bmatrix},\quad
	\bm{\Sigma}_{\xy} =
	\begin{bmatrix}
	\bm{\Sigma}_{\inin} & \bm{\Sigma}_{\inout} \\ 
	\bm{\Sigma}_{\outin} & \bm{\Sigma}_{\outout}\\ 
	\end{bmatrix}.
\end{align}
The mean $\bm{\mu}_{\outgin}$ follows a linear trend on $\bm{x}$ and the scale matrix $\bm{\Sigma}_{\outgin}$ increases as the query point is far from the input marginal distribution, with a dependence on the degree of freedom. % TODO trend on, given ??
These expressions are similar to MVN conditional but the scale ($\bm{\Sigma}_{\outin} = \bm{\Sigma}_{\inin}-\bm{\Sigma}_{\outin}\bm{\Sigma}_{\inin}^{-1}
\bm{\Sigma}_{\outin}^\trsp$ for the MVN) has an additional factor, increasing uncertainty as $\bm{x}$ is far from the known distribution. % TODO maybe redundant with previous sentence

%Gaussian
%\begin{align*}
%%\bm{\mu}_{1|2} &= \bm{\mu}_1 + \bm{\Sigma}_{12}\bm{\Sigma}_{22}^{-1}(\bm{x}_2-\bm{\mu}_2)\\
%\bm{\Sigma}_{\xout|\xin} &= \bm{\Sigma}_{\xout\xout}-\bm{\Sigma}_{\xout\xin}\bm{\Sigma}_{\xin\xin}^{-1}
%\bm{\Sigma}_{\xout\xin}^\trsp
%\end{align*}

\subsection{Bayesian analysis of the mixture model}

Using the conjugate prior for the MVN leads to very efficient training of mixtures with mean-field approximation or Gibbs sampling. % TODO maybe remove mean field, maybe add equation for mixture
Efficient algorithms similar to expectation maximization (EM) can be derived.
For brevity, we will here only summarize the results relevant to our application (see e.g., \cite{bishop2006pattern} for details). % TODO interesting for or to or what ? 
Using mean-field approximation and variational inference, the posterior predictive distribution of a mixture of MVN is a mixture of multivariate t-distributions
\begin{align}
	p(\xy|\XY) = \sum_{k=1}^{K} \pi_k \
	t_{\nu_k - d + 1}(\xy|\,\bm{\mu}_k, \frac{\bm{T}_k(\kappa_k + 1)}{\kappa_k(\nu_k - d + 1)}).
	\label{equ:posterior_pred}
\end{align}
The conditional posterior distribution is then also a mixture
\begin{align}
%p(\bm{x}_{\xout}|\bm{x}_\xin,\bm{X}) &= \sum_{k=1}^{K} p(k|\, \bm{x}_\xin, \bm{X}) p(\bm{x}_\xin|\bm{x}_\xout, k, \bm{X})
%%%%%%%%%%%%%%%%%
p(\bm{y}|\,\bm{x},\bm{X}, \bm{Y}) &= \sum_{k=1}^{K} p(k|\, \bm{x}, \bm{X}, \bm{Y}) p(\bm{y}|\,k, \bm{x},  \bm{X}, \bm{Y}),
%&= \sum_{k=1}^{K} \frac{
%	\pi_k t_{\nu_{\xin, k}}(\bm{x}_\xin|,\bm{\mu}_{\xin, k}, \bm{\Sigma}_{\xin\xin, k})
%}{
%	\sum_{j} \pi_j t_{\nu_{\xin, j}}(\bm{x}_\xin|,\bm{\mu}_{\xin, j}, \bm{\Sigma}_{\xin\xin, j})
%}\
%t_{
%	\nu_{\xout|\xin, k}}(\bm{x}_\xout|,\bm{\mu}_{\xout|\xin, k}, \bm{\Sigma}_{\xout|\xin, k})\\
	\label{equ:posterior_cond_pred}
\end{align}
where we need to compute the marginal probability of the component $k$ given the input $\bm{x}$
\begin{align}
% p(k|\, \bm{x}_\xin, \bm{X})	&=\frac{
%		\pi_k t_{\nu_{\xin, k}}(\bm{x}_\xin|,\bm{\mu}_{\xin, k}, \bm{\Sigma}_{\xin\xin, k})
%	}{
%		\sum_{j} \pi_j t_{\nu_{\xin, j}}(\bm{x}_\xin|,\bm{\mu}_{\xin, j}, \bm{\Sigma}_{\xin\xin, j})
%	}
%%%%%%%%%%%%%%%%%%%
p(k|\, \bm{x}, \bm{X})	&=\frac{
	\pi_k t_{\nu_{\inin, k}}(\bm{x}|\,\bm{\mu}_{\inin, k}, \bm{\Sigma}_{\inin, k})
}{
	\sum_{j} \pi_j t_{\nu_{\inin, j}}(\bm{x}|\,\bm{\mu}_{\inin, j}, \bm{\Sigma}_{\inin, j})
}
	\label{equ:marginal_prob_k}
\end{align}
and apply conditioning in each component using \eqref{mtd_condition_nu}--\eqref{mtd_condition_scale}
\begin{align}
%p(\bm{x}_\xin|\bm{x}_\xout, k, \bm{X}) &=	t_{
%		\nu_{\xout|\xin, k}}(\bm{x}_\xout|,\bm{\mu}_{\xout|\xin, k}, \bm{\Sigma}_{\xout|\xin, k}).
%%%%%%%%%%%%%%%%%%%%%
	p(\bm{y}|k, \bm{x}, \bm{X}, \bm{Y}) &=	t_{
		\nu_{\bm{x}|\bm{y}, k}}(\bm{y}|\,\bm{\mu}_{\bm{y}|\bm{x}, k}, \bm{\Sigma}_{\bm{y}|\bm{x}, k}).
\end{align}
Equation \eqref{equ:marginal_prob_k} exploits the property that marginals of multivariate-t distributions are of the same family \cite{roth2012multivariate}.

\begin{figure}
	\centering
	\includegraphics[width=1.0\linewidth]{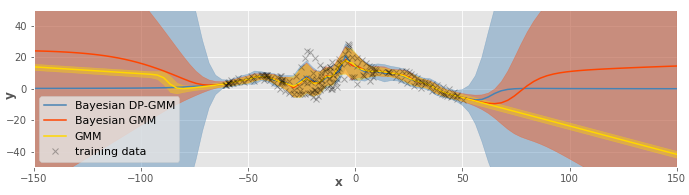}
	\caption{Conditional distribution $p(\bm{y}|\bm{x})$ where $p(\bm{x}, \bm{y})$ is modeled as a mixture of Gaussians. The mean and the standard deviation are represented by applying moment matching on the multimodal conditional distribution. \textit{(blue)} The joint distribution has a Dirichlet process prior. \textit{(red)} The prior is a Dirichlet distribution, with a fixed number of clusters. \textit{(yellow)} The prediction is done without integrating the posterior distribution, which results in a lack of estimation of its uncertainty far from training data.}
	\label{fig:gmmgmrdpcomp}
\end{figure}

\paragraph{Dirichlet distribution or Dirichlet process}
% TODO add a graph to show the difference, notebook already done
% TODO DONE REVIEW : sensitivity is greater with DP
In \cite{bishop2006pattern}, the Bayesian Gaussian mixture model is presented with a Dirichlet distribution prior over the mixing coefficients $\bm{\pi}$. 
An alternative is to use a Dirichlet process, a non-parametric prior with an infinite number of clusters.
For the learning part, this allows the model to cope with an increasing number of datapoints and adapting the model complexity.
Very efficient online learning strategies \cite{hughes2013memoized} exist. % TODO on-line ? ; with previous sentence
With a Dirichlet process, the posterior predictive distribution is similar to \eqref{equ:posterior_pred} but with an additional mixture component being the prior predictive distribution of the MVN. % TODO maybe talk about weight being dependent on the $\alpha$
Conditional distribution given points far from the training data would have the shape of the marginal predictive prior, similarly as in GPR. % DONE - TODO add as shown in Fig
Fig.~\ref{fig:gmmgmrdpcomp} illustrates the difference, when conditioning, between the Dirichlet process and the distribution.
When applied to a policy, it means that when diverging from the distribution of visited states, the control command distribution will match the given prior. 

With a Dirichlet process, the model presented here is a particular case of \cite{hannah2011dirichlet}, having the advantages of faster training with variational techniques and faster retrieval with closed-form integration. 

When using a Dirichlet process prior, the number of clusters is determined by the data. This number is particularly influenced by the hyperparameters $\nu$ and $\bm{T}$ of the Wishart prior. A high $\nu$ leads to a high number of small clusters. On the contrary, a high $\bm{T}$ implies an important regularization on the covariances, which leads to a small number of big clusters. In some cases, it is easier to determine a fixed number of clusters with a Dirichlet distribution. In that case, $\nu$ and $\bm{T}$ can be set small, such that the posterior is more influenced by the data than by the prior.
% TODO DONE REVIEW : how to select priors, what is the sensitivity, maybe add a paragraph at the end
%It seems that conditional posterior predictive distribution Sec.
%III.A.d) depends on the choice of the free parameters T,\nu and
%\kappa, first mentioned in Sec. III.A.a).
%How much sensitive is the method to those parameters? Where they hard
%to tune? Could the authors give an insight about this?

%Online learning , life long, more pessimistic
%keep care of prior

%stochastic online variational learning \cite{hoffman2013stochastic}, sufficient statistics \cite{hughes2013memoized}
%
%\subsection{Benefits of Bayesian GMM conditioning}
%%Neural network no uncertainty. Bayesian yes but with higher cost. Is efficient Our model alleviate
%There are numerous regression techniques.
%Robotic LWPR, Gaussian process, multimodality
%A lot of interesting features come at a very low computation cost and with very few parameters to tune.
%
%\section{Stabilizing distributions of policy} % fusion, product
\section{Product of policy distributions} % fusion, product
%\section{Two heads are better than one: controller fusion} % fusion, product
\label{sec:policy_fusion}
% TODO maybe cite some other works with PoG if some place left
In cognitive science, it is known that two heads are better than one if they can provide an evaluation of their uncertainty when bringing their knowledge together \cite{bahrami2010optimally}. % TODO or neuroscience capital letter in neuroscience ?, split in two sentences
In machine learning, fusing multiples sources is referred to as products of experts (PoE) \cite{hinton1999products}. % TODO check capital letters
In this section, we propose to exploit the uncertainty presented in the above, by fusing multiple policies $p_1(\bm{u}|\bm{x}), \dots, p_{\nbprod}(\bm{u}|\bm{x})$, coming from multiple sources or learning strategies. % TODO check if bracket around list of policies

% TODO DONE REVIEW ; rewrite
%Sec. IV, 2nd paragraph: This paragraph is hard to follow, and not very
%clear. Here is a suggestion on how to rewrite it:
%
%In the general case, computing the mode of a PoE requires optimization.
%Analytical approximations of the posterior can be computationally
%intensive, and inappropriate in applications where the policy should be
%computed fast. However, when assuming a MVN for each expert i,N_i(...),
%the computation becomes lightweight, since the product distribution has
%the closed form expression...

In the general case, computing the mode of a PoE requires optimization, which is inappropriate in applications where the policy should be computed fast.
However, when assuming an MVN for each expert $i$, $\mathcal{N}_i(\bm{x}|\, \bm{\mu}_i, \bm{\Lambda}^{-1}_{i})$, the computation becomes lightweight. The product distribution has the closed form expression
\begin{align}
%	\mathcal{N}_1(\bm{x}|\, \bm{\mu}_1, \bm{\Lambda}^{-1}_{1}),\,
%	\mathcal{N}_2(\bm{x}|\, \bm{\mu}_2, \bm{\Lambda}^{-1}_{2})\\
%	\bm{\Lambda}_p = \sum_{i=1}^{\nbprod} \bm{\Lambda}_i,\ \bm{\mu}_p = \bm{\Lambda}_p^{-1}\sum_{i=1}^{\nbprod} \bm{\Lambda}_i \bm{\mu}_i\\
	\bar{\bm{\Lambda}} = \sum_{i=1}^{\nbprod} \bm{\Lambda}_i,\quad \bar{\bm{\mu}} = \bar{\bm{\Lambda}}^{-1}\sum_{i=1}^{\nbprod} \bm{\Lambda}_i \bm{\mu}_i,
%	\bm{\Lambda} = \sum_{i=1}^{\nbprod} \bm{\Lambda}_i,\ \bm{\mu} = \bm{\Lambda}^{-1}\sum_{i=1}^{\nbprod} \bm{\Lambda}_i \bm{\mu}_i\\
	\label{equ:pog}
\end{align}
where $\bm{\Lambda}$ denotes the precision matrix (inverse of covariance).
This result has an intuitive interpretation: the estimate is an average of the sources weighted by their precisions. % TODO their means or mean ? same for precision
% TODO REVIEW explain better, reformulate
%"using an expert being the Bayesian GMM conditioning" I don't
%understand what this means

If we want to fuse the policy imitation that we proposed in \eqref{equ:posterior_cond_pred} with another policy, several alternatives are possible. If speed is a priority, for example when using torque control, the full conditional distribution \eqref{equ:posterior_cond_pred} could be approximated as an MVN, which enables fast fusion using \eqref{equ:pog}. For this purpose, moment matching can be used. This approximation can be harmful in case of clearly multimodal policies, as illustrated in Fig.~\ref{fig:mmvsmm}\textit{(b)}. If more time is available for computation and/or a higher precision is required, \eqref{equ:posterior_cond_pred} can be estimated as a mixture of MVN by applying moment matching to each cluster. The product of a mixture of MVN and another mixture (or just an MVN) is also a mixture of MVN \cite{gales2006product}. 
To give an idea about computation time, the product between two mixtures of $K=25$ components and $d_u=7$ dimensionality of $\bm{u}$ takes about 3 ms on a standard computer with NumPy. The product between this mixture and an MVN takes about 0.25 ms.
In the experiments presented in Sec.~\ref{sec:experiments}, the global moment matching approximation with an MVN is used, as the tasks do not exhibit multimodality.
 
% TODO rm conditional ? a MVN or an MVN
% TODO add an explanation for the slow one that our imitation policy will have less impact on the product when unsure

% TODO DONE REVIEW : explain that we used MM because our experiments did not really have multimodal policy and we wanted to show a

% TODO for mixture of Gaussians Product [1] M. J. F. Gales and S. S. Airey, “Product of Gaussians for speech recognition,” Comput. Speech Lang., vol. 20, no. 1, pp. 22–40, Jan. 2006.
% easy to interpret example due to lack of space. Add how to do this with mixture (product of each components), maybe make an illustration

%In Sec. IV, the authors propose to approximate the effect of all
%control strategies using the product of experts concept (PoE), which
%ultimately approximates the mixture of multivariate t-distributions by
%a MVN distribution, which is unimodal, using moment matching. However,
%in my opinion, this last approximation contradicts one of the main
%motivations of this work, which is the use of multimodal control
%policies, as stated in Sec. III.a). Therein, the authors defend that
%some policies can clearly be multimodal, and that they intend to have a
%method for multimodal policies, which shall improve others like GPs or
%LWPR. I have some questions about this:
%
%1) Is this a contradiction? Could the authors provide a clear
%explanation of how this does not contradict the initial motivation
%given in Sec. III.a)?
%2) How harmful is the approximation of the PoE? Could the authors
%briefly mention something about this?

\subsection{Examples of controllers}
We present a set of policies that can be combined to increase robustness and that will be used in the experiments. 

\paragraph{Optimal control}
If the task can be formulated as a cost (e.g., attaining a given state), an interesting strategy is to use optimal control (OC).
Classically, these techniques require an accurate model of the system and are subject to local minima (for example, in an environment with obstacles). In combination with imitation, we propose to use OC with crude model approximations (e.g., without modeling obstacles).

In order to combine the policy, we need the OC solver to retrieve a distribution of commands $p(\bm{u})$. A first way will be to use the solution as the mean of an MVN and fix its precision heuristically, such that it dominates the imitation policy outside of the training data.
A more rigorous way is to use the maximum entropy principle \cite{ziebart2008maximum}, retrieving a near optimal stochastic policy.
However, this technique is much more computationally demanding than the standard OC problem.
When using linear dynamics and quadratic cost, the maximum entropy solution can be retrieved very efficiently \cite{levine2014learning}, using linear quadratic tracking (LQT) \cite{bohner2011linear} as
% TODO DONE REVIEW  ; explain in note I guess
%In Equation 21, if the parameters are a function of time, is this LQR
%or iLQR control?
\begin{align}
p(\bm{u}_t|\bm{x}_t) = \mathcal{N}(-\bm{K}_t \bm{x}_t + \bm{c}_t, \bm{Q}_t^{-1}).
\label{equ:LQR_policy} 
\end{align}
% TODO maybe for non linear iterative LQR \cite{li2004iterative}
%It should also be investigated if uncertainties about the dynamics or parameters of the cost function can be propagated to the policy. % TODO maybe remove because very unsure or completely obvious from some people
%SC: maybe remove the above sentence, because it also breaks the flow of the text 

\begin{figure*}[!h]
	\centering
	\includegraphics[width=0.9\linewidth]{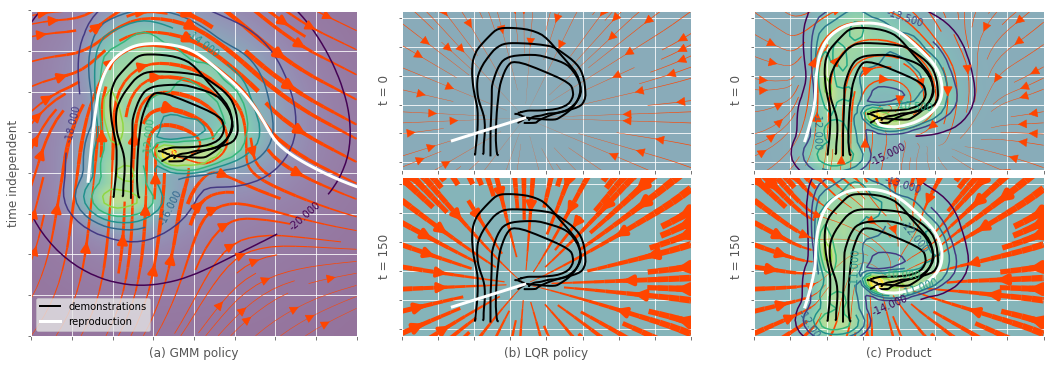}
	\caption{Mixing imitation and LQT, with the final target specified. \textit{(a) Bayesian GMM policy:} Time-independent policy learned with the presented method. \textit{(b) LQT policy:} Only final state target distribution is specified, which results in a time-dependent policy whose control gains increases along time. The shown policies are for $t\in[0, 150]$. \textit{(c) Combination:} The combination is a time-dependent policy. It applies, at the beginning, the imitation policy in zones of high certainty, and is lightly converging outside. At the end of the task, it becomes more strongly converging all over the states. 
}
	\label{fig:lqrproducttime}
\end{figure*}

\paragraph{Time-dependent policy}
For discrete point-to-point tasks, it is possible to use a policy that will ensure convergence at the end of the motion.
This structure is already given in dynamical movement primitives (DMP) \cite{schaal2005learning} or in \cite{hersch2008dynamical}, where a phase variable is used to switch between controllers. 
Our approach allows for a more complex and better-motivated fusion of policies. % TODO maybe remove or check if can be placed before or after.
This stable controller can either be engineered or computed with OC. % TODO change with in this work ?
Here, we used an LQT where the cost on the state is active only at the end of the task.
The controller, in the form of \eqref{equ:LQR_policy}, has gain $\bm{K}_t$ and precision matrix $\bm{Q}_t$ increasing along time, as shown in Fig.~\ref{fig:lqrproducttime}b for starting and final time. % TODO check on the form of, check t\in maybe two sentences and explain more in detail 

\paragraph{Conservative policy} % TODO find a name and acronym :), cautious, shy, stay in the known, afraid of unknown
% TODO DONE REVIEW : maybe put an equation on the one that was used
%Can the authors provide more details about the conservative policy?
We also propose to use a policy that brings us back to the distribution of states where the policy is known, as in \cite{paraschos2015model}. % TODO maybe mv paraschos to later
In their case, the policy was designed as a PD controller, but we propose to use OC or model-based policy search for more generality.
In the proposed GMM, the conservative policy can either be optimized to minimize the uncertainty of the imitation policy (e.g., given by the entropy of the conditional distribution) or to converge to the marginal distribution of states. % TODO maybe rm given by ... of the conditional 
For the experiments, we chose to solve the latter with an LQT. A local quadratic approximation of
$-\log p(\bm{x}|\bm{X})$ from \eqref{equ:posterior_pred} is used as cost on the state. A quadratic cost on the control $\bm{u}$ is set to limit forces or velocities. For an increased precision, this policy can also be learned using maximum entropy model-based policy search \cite{levine2018reinforcement}.

\begin{figure}
	\centering
	\includegraphics[width=.9\linewidth]{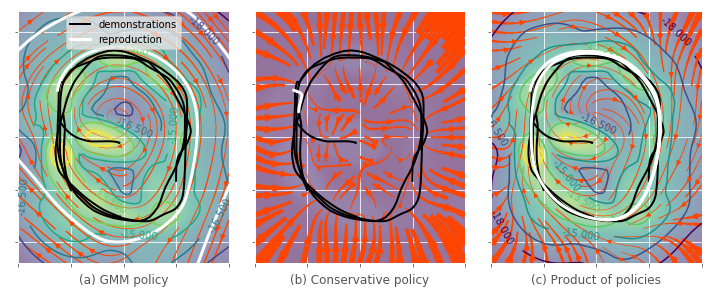}
	\caption{Encoding a limit cycle can be done without any modification of our approach. \textit{(a) Bayesian GMM policy:} By only applying this policy, the robot quickly diverges from the cycle and goes into regions where the policy is unknown. {(b) Conservative policy:} This policy forces the robot to go back to regions where the robot knows the policy, which has a stabilizing function. \textit{(c) Product of policies:} The product of the two policies follows the imitated policy in known regions and converges back to these regions when the robot is brought outside.}
	\label{fig:cyclic}
\end{figure}

Fig.~\ref{fig:cyclic}b illustrates this policy and shows that this technique can encode cyclic motions, which would have been difficult to achieve with the previous propositions. % TODO maybe rm ... to achieve
We also note that the combination given by \eqref{equ:pog} is more complex than the scalar combination proposed in \cite{paraschos2015model} because it can take into account that policies may have variable precisions along different axes. %TODO check note, make aware ?, mv in the experiments maybe

% TODO where to put that our approach is better motivated because takes into account uncertainty blending of policy, better motivated moreover, more than only scalar combination
%activated when outside of the trajectory distribution 

%TODO add code here
Accompanying Python codes and videos are available at \href{https://gitlab.idiap.ch/rli/pbdlib-python}{https://gitlab.idiap.ch/rli/pbdlib-python}.

\section{Experiments}
\label{sec:experiments}
\begin{figure*}[!h]
	\centering
	\includegraphics[width=.9\linewidth]{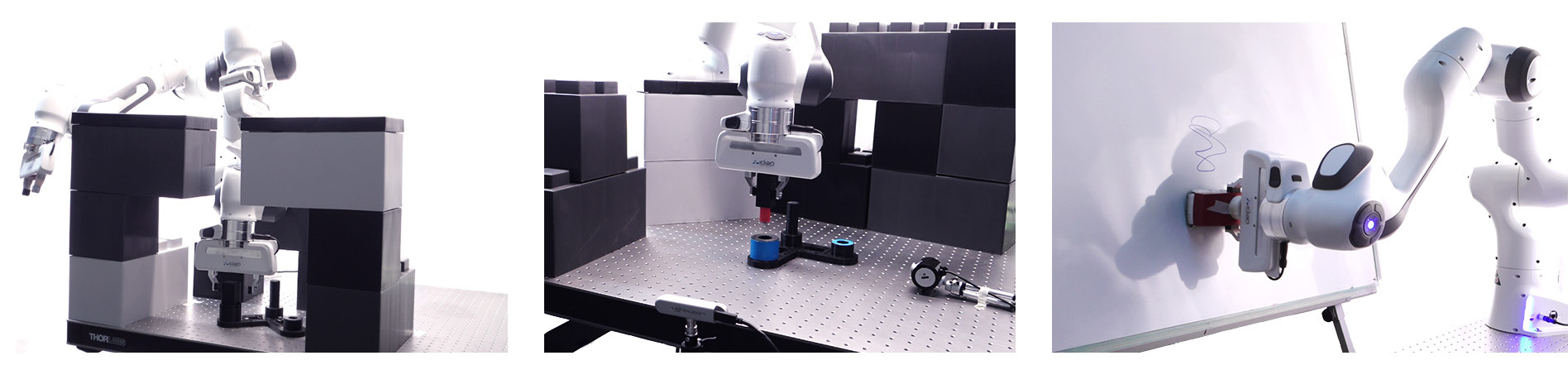}
	\caption{\textit{(left)} Obstacle navigation with joint velocity controller. \textit{(center)} Vision-based peg-in-hole with velocity controlled robot. \textit{(right)} White board wiping with force controlled robot.}
	\label{fig:screenshots}
\end{figure*}

Three experiments are presented to demonstrate that the proposed approach can be used in a variety of tasks. 
They are performed on a 7-axis Panda (Franka Emika) robot.

\subsection{Obstacle navigation}
% insert_obstacle_002.npy
% 13 demonstrations, 5 total of 15s, 8 partial, totaling 115s of demos
In the first experiment, the robot should navigate through fixed obstacles from various initial configurations in order to grasp an object, see Fig.~\ref{fig:screenshots}-\textit{(left)}. 
The system is defined with joint angles $\bm{x} = \bm{q}$ as states and joint velocities
$\bm{u} = \bm{\dot{q}}$ as control commands. % TODO check order
% TODO DONE REVIEW : if time, evaluate without partial demonstrations and with. At least explain that more precision was need at the end
% to precisely align
%
%Can the authors quantify partial demonstrations? Why are these needed?
%how does performance change without these?

In total, 13 demonstrations were recorded (totaling 115 sec of recording). Eight demonstrations were recorded by starting far from the desired point, showing how to avoid the obstacles. The five others were only local, providing more precise demonstrations at the end of the motion, where an accurate alignment between the gripper and the object to grasp is required.
% TODO DONE REVIEW: be more clear that figure 4 is a 2d illustrations and not the real system, just to have a idea
%I do not understand Fig 4. Is this in the end-effector space? But the
%state in this example is the joint angles, and control joint velocity.
%So how is this visualization 2-dimensional?

We chose to fuse the imitation policy with a LQT conservative policy, converging to the marginal distribution of joint angles. This conservative policy is similar to the one illustrated in Fig.~\ref{fig:cyclic} for a 2D system. % TODO maybe change conservative, adapt to decision above
The non-Bayesian policy, as planned, was quickly diverging and dangerously accelerating outside of the demonstrated regions of the state space.

While an optimal control/planning approach would require the modeling of the obstacles and the robot volume, ours was able to propose a solution robust to a wide range of starting points, that could be set up intuitively within only a few minutes. 
% TODO HALF DONE REVIEW : explain the problems, generally be more clear that we are not limited to trajectories
This experiment also illustrates the advantages of learning a distribution of policies instead of trajectories as in \cite{paraschos2013probabilistic}. The teaching time can be optimized with partial demonstrations. Moreover, there is no need for an additional mechanism to realign the demonstrations in time.

\subsection{Vision-based peg-in-hole}
% insert_bicamera_003.npy
% 19 demonstrations of 3.5 +/- 0.7 s, 64 s in total 
In the second task, the robot should insert a peg in a moving hole by looking from two cameras placed on the side, see Fig.~\ref{fig:screenshots}-\textit{(center)}.
The center of the red peg and the center of the blue tube (top part) are extracted by image processing.
The diameter of the peg is 15\% smaller than the hole, such that the task can be solved with an imperfect vision system and without impedance control.
The state is the pixel displacement from the hole and the peg from the two cameras $\bm{x} = \begin{bmatrix}\bm{p}_1^\trsp, \bm{p}_2^\trsp \end{bmatrix}^\trsp$.
The control command is the Cartesian velocity of the gripper in robot frame (the orientation was held fixed) $\bm{u} = \bm{v}\ee$, but joint angle velocities would have been possible as well, likely requiring some additional demonstrations. 

As an evaluation, the robot was initialized at 10 random postures (with the peg still being seen by the two cameras). 
We evaluated its capacity to insert the peg without touching the border of the hole or hitting the support (receptacle). % TODO change support
Multiple combinations of policies were used (among imitation, conservative and optimal control, as presented in Sec.~\ref{sec:policy_fusion}). 
% TODO DONE REVIEW put an equation
% How was the dynamics approximated? Can the authors give some details?
The conservative policy was an LQR, similar to the one in the previous experiment. Instead of converging to the distribution of known joint angles, this policy acts in pixel space.

Computing this policy requires the dynamic model $\bm{x}_{t+1} = f(\bm{x}_t, \bm{u}_t)$, unknown in this experiment. It is learned by recording 10 sec of random motions. The relation between the position of the end-effector $\bm{x}\ee$ and the pixel position of camera $i$, $\bm{p}_i= g_i(\bm{x}\ee)$, is learned using the method presented in Sec.~\ref{sec:gmm}. The Jacobian of $g_i$, which links $\bm{v}\ee$ to pixel velocities, is then used to build a linearized version of $f(\bm{x}_t, \bm{u}_t)$. Due to observation noise, this approach is more robust than directly learning $f(\bm{x}_t, \bm{u}_t)$.

The cost for the optimal control policy is defined as the negative log-likelihood of the distribution of desired final states, encoded as an MVN.
The cost on the state is only active at the end of the LQT planning horizon, corresponding to the length of the longest demonstration. This allows variability during the task and forces convergence at the end of the task.
The cost on $\bm{u}$ is the negative log-likelihood of the distribution of control command $\bm{u}$ during the demonstrations. % The obstacles were not learned. % DONE TODO give details about optimal control
Results are reported in Table~\ref{table:peg_hole}.

\begin{table} % DONE TODO make table
\begin{center}
	% TODO DONE REVIEW : touching without moving the setup
	% What is a slight touch? Can authors quantify this?
	\caption{Results of the peg-in-hole task with different combinations of policies (imitation, optimal control (OC) and conservative (CS)). The task is considered a success if the peg is in the hole without touching the border, or with a slight touch that did not displace the receptacle. Hitting the border or being stuck is considered as a failure.}
   \begin{tabular}{l|ll|l}
	\toprule
%	\multicolumn{3}{c}{Spanning text}     \\
	Policies & \textbf{success rate } & \hspace{-25px} (10 trials) & \textbf{failure }\\
	& without touching & with touching  & \\
	\midrule    
	Imitation only   & 0.3        & 0.1    & 0.6    \\
	Im. + OC    & 0.4        & 0.4       & 0.2   \\
	Im. + CS    & 0.7    & 0.1       & 0.2    \\
	\textbf{Im. + CS + OC}    & \textbf{0.8}    & \textbf{0.1}       & \textbf{0.1}    \\
	OC    & 0.0    & 0.1       & 0.9    \\
	\bottomrule
\end{tabular}
	\end{center}
	\label{table:peg_hole}
\end{table}

Imitation alone shows the effect of an accumulation of errors. It often fails if brought far from the known regions of the state space. It also only slowly stops.
Since the obstacles are neither modeled nor learned, the OC policy results in a straight line motion to the target (as illustrated in Fig.~\ref{fig:lqrproducttime}b) and thus almost always hits the obstacles or the border of the receptacle. It only works when employed very close to the goal. It provides good correction in this context when used in combination with the other policies.
The conservative policy has a relevant effect to converge back to the known regions of the state space and provide a good improvement of the results.

% TODO DONE REVIEW : explain better next paragraph
%I do not understand the last paragraph of this section. is this
%referring to the current experiments, or next section?

In this experiment, we are in a similar situation to \cite{khansari2011learning}, where we control the velocity of an object. But as the relation between end-effector and pixel velocities is not known beforehand and might change, the approach in \cite{khansari2011learning} is not applicable.

\subsection{Force based board wiping}
% TODO DONE REVIEW : add force plot if possible
%Plots showing applied forces on hardware would be very helpful here.

\begin{figure}
	\centering
	\includegraphics[width=.9\linewidth]{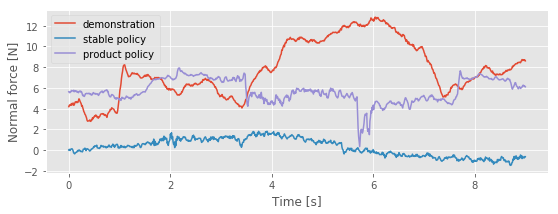}
	\caption{\new{Comparison of normal forces applied by the robot on the board. Using the conservative (or stable) policy results in almost no force. Using the product of policies, the forces are much closer to the recorded ones.}}
	\label{fig:experimentforceplot}
\end{figure}

% clean_001.npy 
% 8 demonstrations totaling 68 s 1 doing 3 turns, 7 how to recover and initiate the turn
In the third task, we demonstrate learning of a force policy within a cyclic motion. % DONE TODO check force policy, bla bla, maybe rm non-discrete (should put somewhere that this task could not be done with trajectory based approaches) put after
The robot has to wipe a board by applying a force against it while doing circular motions. % TODO against or perpendicular
The state is composed of the position and linear velocity of the end-effector (defined at the gripper) in the robot base frame $\bm{x}\! =\! \begin{bmatrix}{\bm{x}\ee}^\trsp, {\bm{v}\ee}^\trsp\end{bmatrix}^\trsp$.
The control command is the force to apply at the end-effector $\bm{u}\! =\!\bm{F}\ee$, which is then transformed to torques with $\bm{\tau} = {\bm{J}(\bm{q})}^\trsp \bm{F}\ee + \bm{c}(\bm{q}, \bm{\dot{q}}) + \bm{g}(\bm{q})$, where Coriolis $\bm{c}$ and gravity $\bm{g}$ compensation torques are added. 
To be able to record forces during demonstrations, bilateral teleoperation is used with a second 7-axis Panda robot. % TODO check teleoperation  Fig.~\ref{fig:cyclic}
The two end-effectors are linked to each other using a virtual spring-damper system with a position offset. % TODO PD controller instead of spring-damper ?
One demonstration of the circular cyclic motion is performed, 7 others are performed showing how to start/recover when not in contact with the board or outside of the wiping area. % DONE TODO check one check number of demonstrations, %TODO maybe project demonstrations on the image
As in the first experiment, we used a combination of the imitation policy and the conservative policy, which forces convergence to the marginal distribution of position and velocity.
By executing only the imitation policy, the robot is pressing against the board, applying a very imprecise force tangentially to the wiping path to compensate for friction. It diverges quickly from the original path. % TODO check order of this sentence and next, how to link them together
The applied forces are quite clumsy (except perpendicularly to the board) and show a wide variance, which explains the failure of this policy.
The conservative policy alone executes circle-shaped motions very robustly without applying any force against the board, resulting in no cleaning, as shown in Fig.~\ref{fig:experimentforceplot}.

Fusion with full precision matrices as in \eqref{equ:pog} allows the imitation policy to be used perpendicularly to the board while using the conservative policy along the other directions, in accordance to their respective precisions. % TODO exhibit instead of show ?, TODO as in or nothing, ADDED according to their ... to show that it is learned
The resulting policy is very robust, even when the user strongly perturbs the robot, showing that it slowly converges back to the board and to the circular motion before starting wiping. % TODO maybe put the emphasis that this is done without changing anything
This robustness and combination of periodic (wiping) and discrete (converging back to the wiping area) would have been very difficult to achieve (if not impossible) with trajectory-based or time-dependent approaches. % DONE TODO check if not impossible

\section{Conclusion}
In this paper, we presented a Bayesian regression technique with several interesting characteristics for robotic applications. % TODO should I ask a sentence to remember everything
We applied this technique to the common problem of distributional shift in policy imitation, where the uncertainty can be exploited for an intelligent fusion of controllers. % TODO general ? well-known ? common ? should I keep alwaysy policy imitation or behavior cloning or mix the two, same for policy (more RL) or controller (more robotic)
Many approaches for learning robotic manipulation tasks impose structure or restrictions on the policy, which limit their range of applications. We showed in three distinct experiments that our approach can be applied without modification to many state-control systems and to a variety of tasks (discrete or/and periodic). % TODO or a mix of the two ? or or/and

Finally, we believe that this method is well suited in an active learning scenario, which would be investigated in future work. In such a case, the robot would ask for demonstrations in unknown regions in order to optimize the learning process. The efficiency of training and the closed-form predictive distribution of the proposed method makes it possible to maximize complex learning criteria.

\bibliography{IEEEabrv,root}

\begin{thebibliography}{10}
\providecommand{\url}[1]{#1}
\csname url@samestyle\endcsname
\providecommand{\newblock}{\relax}
\providecommand{\bibinfo}[2]{#2}
\providecommand{\BIBentrySTDinterwordspacing}{\spaceskip=0pt\relax}
\providecommand{\BIBentryALTinterwordstretchfactor}{4}
\providecommand{\BIBentryALTinterwordspacing}{\spaceskip=\fontdimen2\font plus
\BIBentryALTinterwordstretchfactor\fontdimen3\font minus
  \fontdimen4\font\relax}
\providecommand{\BIBforeignlanguage}[2]{{%
\expandafter\ifx\csname l@#1\endcsname\relax
\typeout{** WARNING: IEEEtran.bst: No hyphenation pattern has been}%
\typeout{** loaded for the language `#1'. Using the pattern for}%
\typeout{** the default language instead.}%
\else
\language=\csname l@#1\endcsname
\fi
#2}}
\providecommand{\BIBdecl}{\relax}
\BIBdecl

\bibitem{ross2011reduction}
S.~Ross, G.~Gordon, and D.~Bagnell, ``A reduction of imitation learning and
  structured prediction to no-regret online learning,'' in \emph{Proc. Intl
  Conf. on Artificial Intelligence and Statistics ({AISTATS})}, 2011, pp.
  627--635.

\bibitem{hersch2008dynamical}
M.~Hersch, F.~Guenter, S.~Calinon, and A.~Billard, ``Dynamical system
  modulation for robot learning via kinesthetic demonstrations,'' \emph{{IEEE}
  Trans. on Robotics}, vol.~24, no.~6, pp. 1463--1467, 2008.

\bibitem{khansari2011learning}
S.~M. Khansari-Zadeh and A.~Billard, ``Learning stable nonlinear dynamical
  systems with gaussian mixture models,'' \emph{{IEEE} Trans. on Robotics},
  vol.~27, no.~5, pp. 943--957, 2011.

\bibitem{paraschos2013probabilistic}
A.~Paraschos, C.~Daniel, J.~R. Peters, and G.~Neumann, ``Probabilistic movement
  primitives,'' in \emph{Advances in {N}eural {I}nformation {P}rocessing
  {S}ystems ({NIPS})}, 2013, pp. 2616--2624.

\bibitem{laskey2017dart}
M.~Laskey, J.~Lee, R.~Fox, A.~D. Dragan, and K.~Y. Goldberg, ``{DART}: Noise
  injection for robust imitation learning,'' in \emph{Conference on {R}obot
  {L}earning ({CoRL})}, 2017, pp. 143--156.

\bibitem{schaal2005learning}
S.~Schaal, J.~Peters, J.~Nakanishi, and A.~Ijspeert, ``Learning movement
  primitives,'' in \emph{Intl Journal of Robotic Research}.\hskip 1em plus
  0.5em minus 0.4em\relax Springer, 2005, pp. 561--572.

\bibitem{calinon2014task}
S.~Calinon, D.~Bruno, and D.~G. Caldwell, ``A task-parameterized probabilistic
  model with minimal intervention control,'' in \emph{Proc.\ {IEEE} Intl Conf.\
  on Robotics and Automation ({ICRA})}, 2014, pp. 3339--3344.

\bibitem{finn2016guided}
C.~Finn, S.~Levine, and P.~Abbeel, ``Guided cost learning: Deep inverse optimal
  control via policy optimization,'' in \emph{Proc.\ Intl Conf.\ on Machine
  Learning ({ICML})}, 2016, pp. 49--58.

\bibitem{rajeswaran2017learning}
A.~Rajeswaran, V.~Kumar, A.~Gupta, G.~Vezzani, J.~Schulman, E.~Todorov, and
  S.~Levine, ``Learning complex dexterous manipulation with deep reinforcement
  learning and demonstrations,'' \emph{Proc.\ Robotics: Science and Systems
  ({RSS})}, 2018.

\bibitem{nair2018overcoming}
A.~Nair, B.~McGrew, M.~Andrychowicz, W.~Zaremba, and P.~Abbeel, ``Overcoming
  exploration in reinforcement learning with demonstrations,'' in \emph{Proc.\
  {IEEE} Intl Conf.\ on Robotics and Automation ({ICRA})}.\hskip 1em plus 0.5em
  minus 0.4em\relax IEEE, 2018, pp. 6292--6299.

\bibitem{sung2004gaussian}
H.~G. Sung, ``{G}aussian mixture regression and classification,'' {P}h{D}
  thesis, Rice University, Houston, Texas, 2004.

\bibitem{schaal2002scalable}
S.~Schaal, C.~G. Atkeson, and S.~Vijayakumar, ``Scalable techniques from
  nonparametric statistics for real time robot learning,'' \emph{Applied
  Intelligence}, vol.~17, no.~1, pp. 49--60, 2002.

\bibitem{rasmussen2004gaussian}
C.~E. Rasmussen, ``{G}aussian processes in machine learning,'' in
  \emph{Advanced lectures on machine learning}.\hskip 1em plus 0.5em minus
  0.4em\relax Springer, 2004, pp. 63--71.

\bibitem{bishop2006pattern}
C.~M. Bishop, \emph{Pattern Recognition and Machine Learning (Information
  Science and Statistics)}.\hskip 1em plus 0.5em minus 0.4em\relax Secaucus,
  NJ, USA: Springer, 2006.

\bibitem{murphy2007conjugate}
K.~P. Murphy, ``Conjugate {B}ayesian analysis of the {G}aussian distribution,''
  University of British Columbia, Tech. Rep., 2007.

\bibitem{roth2012multivariate}
M.~Roth, \emph{On the multivariate t distribution}.\hskip 1em plus 0.5em minus
  0.4em\relax Link{\"o}ping University Electronic Press, 2013.

\bibitem{hughes2013memoized}
M.~C. Hughes and E.~Sudderth, ``Memoized online variational inference for
  dirichlet process mixture models,'' in \emph{Advances in {N}eural
  {I}nformation {P}rocessing {S}ystems ({NIPS})}, 2013, pp. 1133--1141.

\bibitem{hannah2011dirichlet}
L.~A. Hannah, D.~M. Blei, and W.~B. Powell, ``{D}irichlet process mixtures of
  generalized linear models,'' \emph{Journal of Machine Learning Research},
  vol.~12, no. Jun, pp. 1923--1953, 2011.

\bibitem{bahrami2010optimally}
B.~Bahrami, K.~Olsen, P.~E. Latham, A.~Roepstorff, G.~Rees, and C.~D. Frith,
  ``Optimally interacting minds,'' \emph{Science}, vol. 329, no. 5995, pp.
  1081--1085, 2010.

\bibitem{hinton1999products}
G.~E. Hinton, ``Products of experts,'' \emph{Proc. Intl Conf. on Artificial
  Neural Networks ({ICANN})}, pp. 1--6, 1999.

\bibitem{gales2006product}
M.~J.~F. Gales and S.~S. Airey, ``Product of {G}aussians for speech
  recognition,'' \emph{Computer Speech and Language}, vol.~20, no.~1, pp.
  22--40, jan 2006.

\bibitem{ziebart2008maximum}
B.~D. Ziebart, A.~L. Maas, J.~A. Bagnell, and A.~K. Dey, ``Maximum entropy
  inverse reinforcement learning.'' in \emph{Proc.\ {AAAI} Conference on
  Artificial Intelligence}, vol.~8.\hskip 1em plus 0.5em minus 0.4em\relax
  Chicago, IL, USA, 2008, pp. 1433--1438.

\bibitem{levine2014learning}
S.~Levine and P.~Abbeel, ``Learning neural network policies with guided policy
  search under unknown dynamics,'' in \emph{Advances in {N}eural {I}nformation
  {P}rocessing {S}ystems ({NIPS})}, 2014, pp. 1071--1079.

\bibitem{bohner2011linear}
M.~Bohner and N.~Wintz, ``The linear quadratic tracker on time scales,''
  \emph{International Journal of Dynamical Systems and Differential Equations},
  vol.~3, no.~4, pp. 423--447, 2011.

\bibitem{paraschos2015model}
A.~Paraschos, E.~Rueckert, J.~Peters, and G.~Neumann, ``Model-free
  probabilistic movement primitives for physical interaction,'' in \emph{Proc.\
  {IEEE/RSJ} Intl Conf.\ on Intelligent Robots and Systems ({IROS})}.\hskip 1em
  plus 0.5em minus 0.4em\relax IEEE, 2015, pp. 2860--2866.

\bibitem{levine2018reinforcement}
S.~Levine, ``Reinforcement learning and control as probabilistic inference:
  Tutorial and review,'' \emph{arXiv preprint arXiv:1805.00909}, 2018.

\end{thebibliography}
\end{document}